# DCNNs on a Diet: Sampling Strategies for Reducing the Training Set Size

Maya Kabkab, Azadeh Alavi, Rama Chellappa

Center for Automation Research, UMIACS
University of Maryland, College Park
{mayak, azadeh, rama}@umiacs.umd.edu

## Abstract

Large-scale supervised classification algorithms, especially those based on deep convolutional neural networks (DCNNs), require vast amounts of training data to achieve state-of-the-art performance. Decreasing this data requirement would significantly speed up the training process and possibly improve generalization. Motivated by this objective, we consider the task of adaptively finding concise training subsets which will be iteratively presented to the learner. We use convex optimization methods, based on an objective criterion and feedback from the current performance of the classifier, to efficiently identify informative samples to train on. We propose an algorithm to decompose the optimization problem into smaller per-class problems, which can be solved in parallel. We test our approach on standard classification tasks and demonstrate its effectiveness in decreasing the training set size without compromising performance. We also show that our approach can make the classifier more robust in the presence of label noise and class imbalance.

## 1 Introduction

Deep learning has recently shown remarkable performance on many complex classification tasks. Currently, the best performing deep networks have many hidden layers and an extremely large number of trainable parameters, therefore requiring vast amounts of training data [1–3]. This raises the question of whether all this data is really necessary and whether all training samples are equally valuable in the learning process. While it is true that presenting the classifier with enough information is essential to achieving good performance, large training set sizes can be detrimental to generalization performance and invariably need significant training time. Such large training sets can often include redundant or noisy samples which only introduce unnecessary computations and could cause learning bias. In this paper, we address the problem of adaptively finding a smaller subset of training samples which allow fast learning without compromising performance.

This problem, sometimes referred to as *exemplar* or *active selection*, has been studied in the literature. Starting with a given set of labeled examples, active selection aims to identify a subset to use for training, while leveraging information obtained from the classifier trained on previous selections. One simple approach [4] repeatedly presents the same example if the network error exceeds a threshold. In [5], this problem is addressed in the context of feedforward neural networks. The authors propose a sequential method to select one training sample at a time such that, when added to the previous set of examples, it results in the largest decrease in a squared error estimate criterion. A similar objective is considered in [6] based on pattern informativeness – a measure of a sample's influence on the classifier output.

.

A closely related approach is *active learning* which starts with an unlabeled set of examples and sequentially identifies critical samples to label and train on [7–10]. It is shown that a classifier trained on a carefully chosen subset can sometimes outperform one that is trained on all the available data. Furthermore, [11] suggests that guiding a classifier by presenting training samples in an order of increasing difficulty can speed up learning and result in convergence to a better local minimum.

In this paper, we present strategies to make optimal use of available training data by adaptively selecting batches of training samples which will be iteratively presented to the classifier. In contrast with active learning, we assume a fully supervised setting where all training samples are labeled and available a priori. We are interested in incrementally training a deep neural network, using batches of training data carefully selected to meet four criteria: class balance, diversity, representativeness, and classifier uncertainty. The class balance criterion utilizes the a priori knowledge of labels to ensure that all classes are appropriately present in the new training batch. We propose a novel class balancing algorithm which uses immediate feedback from the classifier to allot a subset of training samples to each class based on the average classifier performance on that class. Diversity and representativeness are distance-based measures aiming to reduce redundancy while maximizing the quality of selected samples. Such strategies have been used in active learning [12, 13], subset selection [14, 15], and clustering [16]. Finally, the classifier uncertainty criterion favors samples that the classifier has not yet properly learnt, thus driving it to explore unvisited parts of the input space. We combine the last three criteria and use optimization techniques from [17, 18] to identify a near-optimal batch to train on at every iteration. We apply our methods on the problems of digit and face recognition. Our results indicate that the training set size can be significantly reduced without sacrificing performance.

The rest of the paper is organized as follows. The problem formulation is stated in Sections 2.1 - 2.5 and the proposed solution in Section 2.6. Experimental results, comparing our method to random sampling, are presented in Section 3.

## 2  Problem statement

We assume we are given a fixed classifier architecture, and a set of labeled training data points: $\mathcal{X} = \bigcup_{k=1}^{L} \mathcal{X}_k$, where $\mathcal{X}_k = \{X_{1,k}, X_{2,k}, \ldots, X_{N_k,k}\}$ are the training samples belonging to class $k$, and $L$ is the number of classes. At each time instance $t$, we select a subset $\mathcal{B}^t \subset \mathcal{X}$, such that the classifier (which has previously been trained on $\mathcal{B}^{t-1}$) exhibits good generalization performance when trained on $\mathcal{B}^t$.

To this end, we formulate a criterion for selecting new training examples which serves the following objectives:

(O1) The samples in $\mathcal{B}^t$ must be such that the classifier is *uncertain* about classifying them (or *certain* but *wrong* in its classification).
(O2) $\mathcal{B}^t$ should have a *balanced* selection from all classes.
(O3) $\mathcal{B}^t$ should be sufficiently *diverse*.
(O4) $\mathcal{B}^t$ should be *representative* of $\mathcal{X}$.

We will mathematically formulate each of these objectives in the following sections.

### 2.1  Classifier uncertainty and error

We assume that, at time instance $t$, the classifier produces $L$ outputs for each training sample $X_{i,k}$ from class $k$, denoted by

$$p^t(X_{i,k}) = [p_1^t(X_{i,k}), p_2^t(X_{i,k}), \ldots, p_L^t(X_{i,k})], \tag{1}$$

where $p_l^t(X)$ is interpreted as the classifier's estimate of the probability that $X \in \mathcal{X}$ belongs to class $l$, and satisfies $p_l^t(X) \geq 0 \ \forall l$, and $\sum_{l=1}^{L} p_l^t(X) = 1$. In order to quantify classifier uncertainty and error we define:

$$c^t(X_{i,k}) = -\sum_{l=1}^{L} \left(\beta^t \cdot \mathbb{1}[l = k] + (1 - \beta^t) \cdot p_l^t(X_{i,k})\right) \log p_l^t(X_{i,k}), \tag{2}$$

where $\beta^t \in [0, 1]$ is a chosen parameter. We can interpret $c^t(X_{i,k})$ in two ways. First, $c^t(X_{i,k})$ can be seen as a weighted sum of an error term: $-\sum_{l=1}^{L} \mathbb{1}[l = k] \cdot \log p_l^t(X_{i,k}) = \log p_k^t(X_{i,k})$ and an entropy term: $-\sum_{l=1}^{L} p_l^t(X_{i,k}) \log p_l^t(X_{i,k})$. These two terms correspond to the correctness of the classifier's decision and the uncertainty in this decision, therefore satisfying objective (O1) above. Second, $c^t(X_{i,k})$ can be interpreted as a bootstrapping technique to overcome possible label noise [19], in which case $\beta^t \mathbb{1}[l = k] + (1 - \beta^t)p_l^t(X_{i,k})$ is a weighted "correct label" and $c^t(X_{i,k})$ represents the cross-entropy between $p^t(X_{i,k})$ and this weighted label. $c^t(\cdot)$ being low on a given sample means that the classifier has enough information about this sample. In order to present the classifier with informative samples, we would therefore like to pick samples where $c^t(\cdot)$ is large.

## 2.2 Class balance

At each time instance $t$, we would like to select a total of $M^t$ samples, distributed among all classes in a balanced way. However, it might be counter-intuitive to simply impose that all classes be equally represented in the subset $\mathcal{B}^t$, as the current classifier may be performing very well on some of them. Therefore, we assign a budget $M_k^t$ to each class depending on the average performance on this class. This can be measured by $c_k^t = \frac{1}{N_k} \sum_{i=1}^{N_k} c^t(X_{i,k})$, where $c^t(X_{i,k})$ is defined in equation (2). The larger $c_k^t$ is, the more samples we assign to class $k$. An objective function of $\sum_{k=1}^{L} c_k^t \cdot M_k^t$ would result in the trivial solution of assigning all the budget $M^t$ to the class with the largest uncertainty score, and would contradict the class balancing requirement. We therefore use a logarithmic objective function and formulate the problem as follows:

$$\max_{M_k^t \in \mathbb{Z}^+} \sum_{k=1}^{L} \log\left(1 + \alpha \cdot c_k^t \frac{M_k^t}{M^t}\right) \quad \text{s. t.} \quad \sum_{k=1}^{L} M_k^t \leq M^t; \; M_k^t \leq |\mathcal{X}_k|, \tag{3}$$

where $\alpha > 1$ sets the sensitivity of the method (the smaller $\alpha$, the larger the effect of differences in $c_k^t$). This problem arises in information theory, in allocating power to a set of communication channels [20, Section 9.4]. We use a similar formulation since $M_k^t$ represents the budget allocated to the $k^{\text{th}}$ class (channel), and $1/c_k^t$ is akin to channel quality. There exists a very efficient solution to this convex optimization problem, known as the water-filling algorithm [21, Section 5.5], where we interpret water levels as the number of samples allocated to each class. Our formulation differs from the standard formulation due to the addition of the last constraint (which ensures that we do not allocate more samples than available in the pool $\mathcal{X}_k$). Another difference is that the feasible set in (3) is the set of non-negative integers.

**Theorem 1** *The modified water-filling problem in (3) can be solved using Algorithm 1.*

---
**Algorithm 1** Integer water-filling algorithm with caps
---
1: Sort the base levels $\frac{M^t}{\alpha c_k^t}$ in ascending order and take the ceiling of the base levels $\lceil \frac{M^t}{\alpha c_k^t} \rceil$.
2: Place "caps" at $\lceil \frac{M^t}{\alpha c_k^t} \rceil + |\mathcal{X}_k|$.
3: **repeat**
4:     Fill with one water unit at a time proceeding from left to right without exceeding any cap.
5: **until** $M^t$ water units are used or all empty spaces are filled.

---

"Caps" enforce the $M_k^t \leq |\mathcal{X}_k|$ constraints. Each water unit corresponds to one training sample being assigned to a class. An illustration of the algorithm is found in Figure 1 for a budget $M^t = 10$. The numbers on the water units show the order in which they have been assigned. Because of the balanced selection of budgets $\{M_k^t\}$, this formulation addresses the class balance objective (O2).

**Remark 1** *Objectives (O3) and (O4) are only meaningful when considered as intraclass rather than globally. Two images from different classes trivially meet the diversity criterion but cannot be representative of each other. Since we are considering supervised learning settings, we can leverage the label information and focus on finding a diverse representative subset of each class separately. The budget selection algorithm in Theorem 1 allows us to do so by distributing our original budget $M^t$ amongst the various classes. We can therefore solve $L$ independent problems. We drop the class*

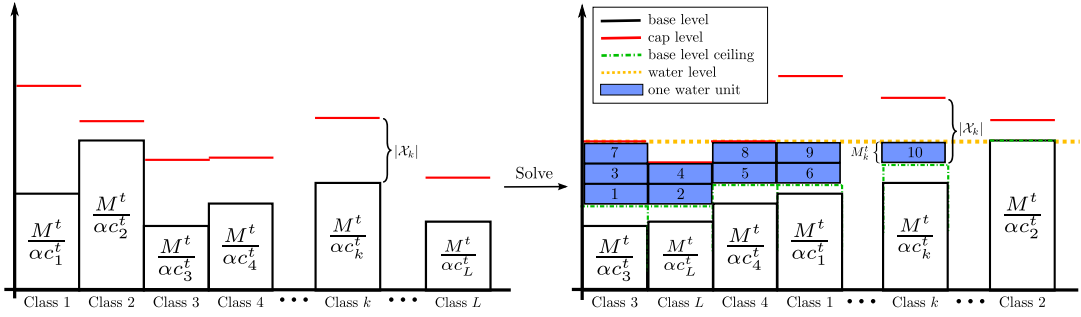

Figure 1: Integer water-filling with caps.

*subscript $k$ and assume that we would like to select a subset $\mathcal{B}^t$ from a pool of samples $\mathcal{X}$, where all the samples belong to the same class. For notational convenience, we also drop the time superscript $t$, with the understanding that this procedure will be repeated at every time step.*

### 2.3 Subset diversity

As per (O3), we would like to select a *diverse* subset, i.e., one that does not have too much redundancy. To this end, we assume we have a distance metric $d(\cdot, \cdot)$ such that $d(X_i, X_j)$ represents the distance between samples $X_i$ and $X_j$. This can, for example, be the Euclidean distance between $X_i$ and $X_j$, or the Euclidean distance between their feature vectors, in some pre-defined feature space. In order to maximize diversity, we seek to **maximize** the average distance between all selected samples, i.e., find $\mathcal{B}$ such that:

$$\frac{1}{M^2} \sum_{X \in \mathcal{B}} \sum_{X' \in \mathcal{B}} d(X, X') \tag{4}$$

is maximized,[1] where $M$ is the budget allocated by the water-filling algorithm. Let $N = |\mathcal{X}|$, the training set size of the class under consideration. We introduce a binary variable $\mathbf{s} \in \{0, 1\}^N$, such that $s_i = 1$ if $X_i \in \mathcal{B}$, and $s_i = 0$ otherwise. We also group all the distances in a matrix $\mathbf{D} \in \mathbb{R}^{N \times N}$ such that $D_{ij} = d(X_i, X_j)$. As such, the objective can be re-written as

$$\max_{\mathbf{s} \in \{0,1\}^N} \frac{1}{M^2} \mathbf{s}^\intercal \mathbf{D} \mathbf{s}. \tag{5}$$

This problem formulation ensures that the chosen samples are sufficiently distant from each other.

### 2.4 Subset representativeness

Per (O4), we would also like to select a *representative* subset $\mathcal{B}$, i.e., the non-selected samples must be well represented by the set $\mathcal{B}$. To this end, we seek to **minimize** the average distance between selected and non-selected samples. As before, this can be re-written as

$$\min_{\mathbf{s} \in \{0,1\}^N} \frac{1}{M(N-M)} (\mathbf{1} - \mathbf{s})^\intercal \mathbf{D} \mathbf{s}, \tag{6}$$

where $\mathbf{1}$ is the vector of all ones in $\mathbb{R}^N$.

### 2.5 Joint formulation

As previously mentioned, once the sub-problem of allocating budgets to each class has been solved, we seek to solve $L$ independent problems of finding a diverse, representative subset over which the classifier performs poorly. We therefore combine the subset diversity, representativeness, and uncertainty criteria. We define the vector $\mathbf{c} \triangleq [c(X_1), \ldots, c(X_N)]^\intercal$ where $c(\cdot)$ is as defined in

---

[1] Other objective functions can be formulated such as maximizing the minimum distance between selected samples. While guaranteeing less redundancy, such objective functions are more difficult to solve.

Section 2.1. To make the quantities comparable, we normalize $\mathbf{D}$ and $\mathbf{c}$ such that all their elements lie in $[0, 1]$. We denote the normalized quantities by $\tilde{\mathbf{D}}$ and $\tilde{\mathbf{c}}$ respectively. Our objective function is:

$$\max_{\mathbf{s}\in\{0,1\}^N} \lambda_1 \cdot \underbrace{\frac{1}{M^2}\mathbf{s}^\mathsf{T}\tilde{\mathbf{D}}\mathbf{s}}_{\text{diversity}} - \lambda_2 \cdot \underbrace{\frac{1}{M(N-M)}(\mathbf{1}-\mathbf{s})^\mathsf{T}\tilde{\mathbf{D}}\mathbf{s}}_{\text{representativeness}} + \lambda_3 \cdot \underbrace{\frac{1}{M}\tilde{\mathbf{c}}^\mathsf{T}\mathbf{s}}_{\text{classifier uncertainty}} , \quad (7)$$

where $\lambda_1, \lambda_2, \lambda_3 \geq 0$ are parameters which dictate the relative importance of each criterion. We need to add the constraint that $|\mathcal{B}| = M$, where $M$ is the budget allocated by the water-filling algorithm. The joint optimization problem for each class is therefore:

$$\min_{\mathbf{s}\in\{0,1\}^N} -\lambda_1 \cdot \frac{1}{M}\mathbf{s}^\mathsf{T}\tilde{\mathbf{D}}\mathbf{s} + \lambda_2 \cdot \frac{1}{N-M}(\mathbf{1}-\mathbf{s})^\mathsf{T}\tilde{\mathbf{D}}\mathbf{s} - \lambda_3 \cdot \tilde{\mathbf{c}}^\mathsf{T}\mathbf{s} \quad \text{s.t.} \quad \mathbf{1}^\mathsf{T}\mathbf{s} = M. \quad (8)$$

It is important to note that the division of our problem into $L$ independent sub-problems provides many advantages. First, formulating the problem on the entire training dataset would require a very large distance matrix $\mathbf{D}$ which would, in most cases, need excessive storage. Second, the $L$ sub-problems are completely independent and can run in parallel, thus reducing computation time.

### 2.6 Proposed solution

The problem in (8) is not convex for two reasons: (i) the set $\{0, 1\}^N$ is finite and therefore not convex, and (ii) the objective function is generally not convex. We change the constraint $\mathbf{1}^\mathsf{T}\mathbf{s} = M$ to its equivalent $(\mathbf{1}^\mathsf{T}\mathbf{s} - M)^2 = 0$ (as this guarantees zero duality gap [17]) and make the change of variable $\mathbf{x} = 2\mathbf{s} - \mathbf{1}$, where $\mathbf{x} \in \{-1, 1\}^N$. Let

$$\mathbf{A} \triangleq -\left(\frac{\lambda_1}{4M} + \frac{\lambda_2}{4(N-M)}\right)\tilde{\mathbf{D}}, \qquad \mathbf{b} \triangleq -\frac{\lambda_1}{2M}\tilde{\mathbf{D}}^\mathsf{T}\mathbf{1} - \frac{\lambda_3}{2}\tilde{\mathbf{c}}. \quad (9)$$

An equivalent problem to (8) is given by:

$$\min_{\mathbf{x}\in\{-1,1\}^N} \mathbf{x}^\mathsf{T}\mathbf{A}\mathbf{x} + \mathbf{b}^\mathsf{T}\mathbf{x} \quad \text{s.t.} \quad (\mathbf{1}^\mathsf{T}\mathbf{x} - 2M + N)^2 = 0. \quad (10)$$

This problem is known as *constrained binary quadratic programming* and is NP-hard [17]. We seek an efficient relaxation to this problem.

**Theorem 2** *The solution $\mathbf{x}^*$ to (10) can be well-approximated by*

$$\hat{\mathbf{x}}^* = -\frac{1}{2}\left(\mathbf{A} + \mu^*\mathbf{1}\mathbf{1}^\mathsf{T} + \gamma^*\mathbf{I}\right)^\dagger (\mathbf{b} - 2\mu^*(2M - N)\mathbf{1}), \quad (11)$$

*where $(\cdot)^\dagger$ denotes the pseudo-inverse, $\mathbf{I}$ denotes the identity matrix in $\mathbb{R}^{N\times N}$, and $\mu^*, \gamma^*$ are the solution to the following semi-definite program (SDP):*

$$\max_{\mu,\gamma,\tau\in\mathbb{R}} \quad (2M-N)^2\mu - \gamma N - \tau$$

$$\text{s.t.} \quad \begin{pmatrix} \tau & \frac{1}{2}(\mathbf{b} - 2\mu(2M-N)\mathbf{1})^\mathsf{T} \\ \frac{1}{2}(\mathbf{b} - 2\mu(2M-N)\mathbf{1}) & \mathbf{A} + \mu\mathbf{1}\mathbf{1}^\mathsf{T} + \gamma\mathbf{I} \end{pmatrix} \succeq 0 \quad (12)$$

We select the samples corresponding to the largest $M$ entries in $\hat{\mathbf{x}}^*$.

## 3 Experiments

In this section, we test the proposed method on several real-world classification tasks. We consider digit and face recognition problems. We compare our approach to the random selection of training samples as used in ordinary training algorithms. Our formulation does not assume a specific classifier structure. However, we will illustrate our results on deep neural networks as they are the current state-of-the-art. We use the Caffe framework [22] for the implementation of Convolutional Neural Networks (CNN) as well as the SDPA framework [23] to solve the SDP problem in (12). We also calculate distances between samples based on the Local Binary Patterns (LBP) features [24].

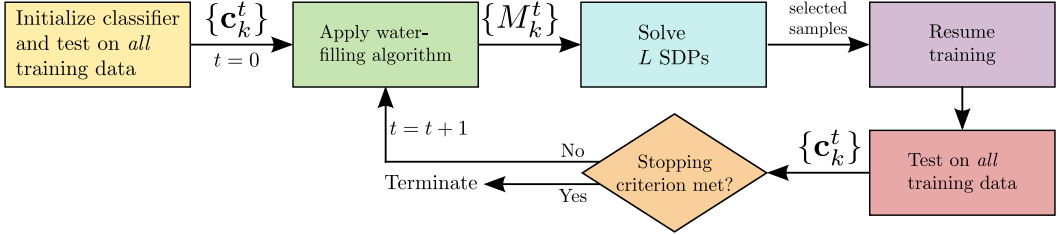

Figure 2: Proposed algorithm.

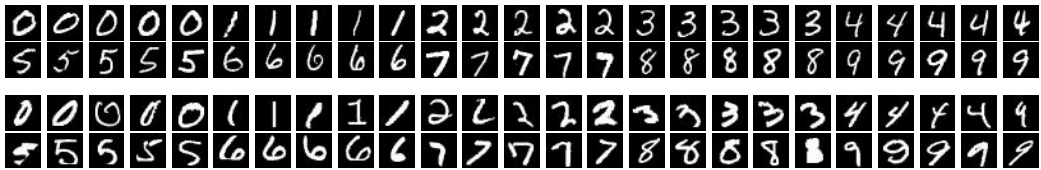

Figure 3: Top: Selected samples when $\lambda_2 = 20\lambda_1$. Bottom: Selected samples when $\lambda_2 = \lambda_1$.

For each of these experiments, unless otherwise specified, we start from a randomly initialized CNN with a fixed architecture. First, we test this CNN on the entire pool of training examples to get the initial average uncertainty levels $\{c_k^0\}_{k=1}^L$. Then, at every time step $t$, we use $\{c_k^t\}_{k=1}^L$ to obtain a class-specific budget using Algorithm 1 and solve (11) and (12) independently for each class, resulting in a new selected batch. We resume training, starting from the previous classifier weights, on the union of all selected batches. At each time step, all candidate samples have a chance to be selected, i.e., previously selected examples are not removed from the set of candidates. We iterate until a stopping criterion is met. We choose this stopping criterion to be a threshold on the classifier error on the entire pool of training samples. As most samples in this pool have not been presented to the classifier, this is an adequate estimate of the generalization performance. The overall algorithm is illustrated in Figure 2.

We do not employ any type of data pre-processing or augmentation techniques which are widely used to achieve state-of-the-art performance, since these methods are not the focus of this work. Instead, we choose to focus on the effect of our training set selection method on the generalization performance, compared to picking the training samples in a random fashion. As our training is incremental, we add dropout layers [25] whenever necessary to combat the problem of catastrophic forgetting in deep neural networks. Catastrophic forgetting refers to the inability of a learning method to preserve previously learnt information when exclusively trained on new data [26, 27].

### 3.1 MNIST digit recognition

For the problem of digit recognition on the well-known MNIST dataset, we use the LeNet architecture [28]. We run our experiments using a randomly selected subset of the MNIST dataset consisting of 1000 images from each class. We use a total budget of 50 training images per one loop of our algorithm (see Figure 2).

#### 3.1.1 Diversity vs. representativeness

We first illustrate the effect of the weights $\lambda_1, \lambda_2$, defined in (7), on the selection process. We set $\lambda_3 = 0$. Figure 3 shows the selected samples when $\lambda_2 = 20\lambda_1$ (top) and $\lambda_2 = \lambda_1$ (bottom). When $\lambda_2$ is large, more representative samples are chosen, as seen in Figure 3, top. When $\lambda_1 = \lambda_2$, more diverse samples are chosen. This validates our initial objective formulation in (7).

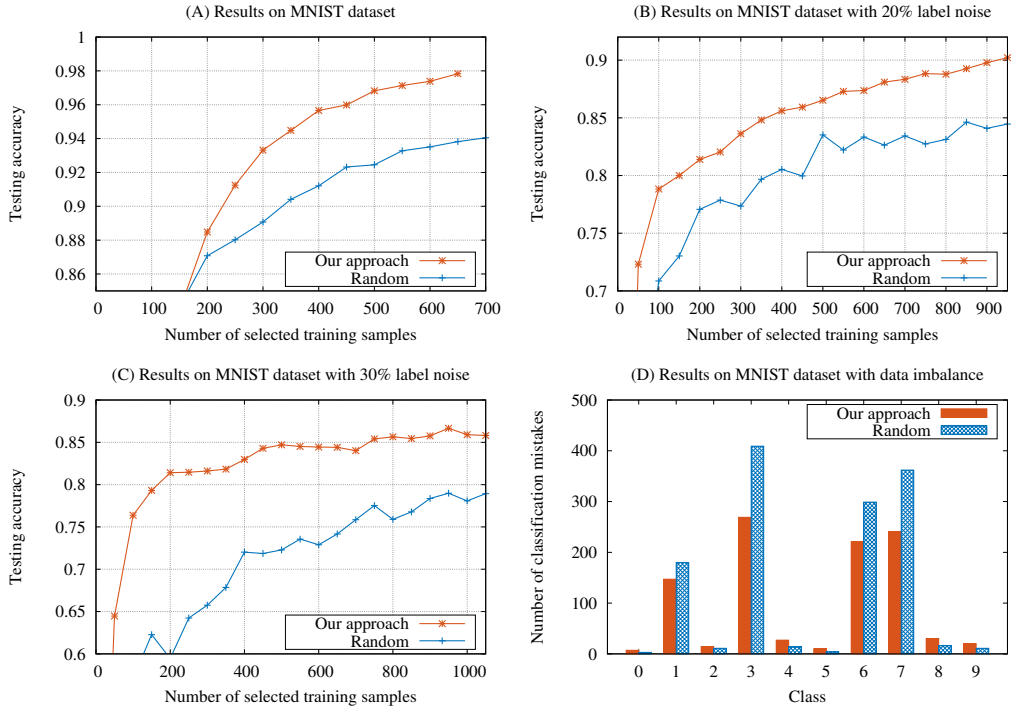

Figure 4: Results on MNIST dataset.

### 3.1.2 Clean labels

We compare our method of adaptively selecting training batches to the baseline of random selection. As discussed in [11], we introduce "easier" samples first and gradually increase the difficulty. We achieve this by keeping $\lambda_1$ fixed, and starting with $\lambda_2 = 10\lambda_1$ and $\lambda_3 = 0$. Picking a large $\lambda_2$ puts more weight on the representativeness term in (7) and thus ensures that outliers are not picked. We gradually decrease $\lambda_2$ and increase $\lambda_3$ in order to allow for more difficult examples to be sampled. We present our findings in Figure 4(A). Our approach outperforms random selection by a margin of $4\%$. Furthermore, the number of samples required by our proposed method to reach a target performance level is much smaller than random sampling. For instance, for a target testing accuracy of $94\%$, around 700 samples are needed for random as opposed to less than 350 samples for our approach.

To assess the quality of the achieved local minimum obtained with our method, we train a CNN from scratch on all of the selected samples introduced at the same time (using the regular non-adapative CNN training). This results in a testing accuracy of $96.3\%$, inferior to the one obtained by our method ($98\%$). This validates the claim that adaptive selection of training data guides the neural network towards a better local optimum. Our algorithm has selected easier training samples in the first few iterations, and more difficult samples later on, as dictated by the change in weights $\lambda_1, \lambda_2, \lambda_3$.

### 3.1.3 Noisy labels

We now assess the performance of our algorithm in the presence of label noise. We randomly change the correct labels in $20\%$ and $30\%$ of the training samples. The results are shown in Figures 4(B) and 4(C), respectively. It is seen that our approach out-performs random selection by more than $5\%$. To combat label noise, we decrease the diversity weight $\lambda_1$ and adopt a more "cautious" approach by increasing $\lambda_3$ at a slower pace. This results in a slower but safer update of the network. In fact, the total number of noisy training images chosen by our algorithm for the case of $20\%$ label noise is 93 images by the $12^{th}$ loop (or $6.5\%$ of the picked images), whereas random sampling obviously picks around $20\%$ noisy samples.

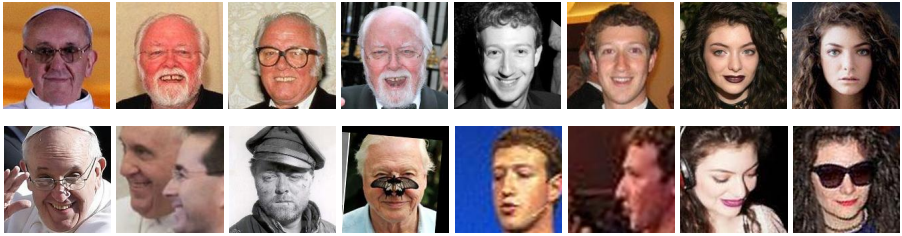

Figure 5: Examples of selected samples at loop 1 (top) and loop 13 (bottom).

### 3.1.4 Data imbalance

Finally, we test our method on a scenario where there is a significant data imbalance between different classes. This can happen when acquiring labeled data for some classes is considerably more difficult than for others. We artificially introduce data imbalance by picking 4 classes at random and reducing their training set size to between 10 and 20 images per class. Our approach achieves $90.14\%$ testing accuracy after only 9 loops of the algorithm (i.e., 450 picked samples), while random sampling achieves $86.91\%$ using the entire training set. We are thus able to boost the performance by over $3\%$ while only using a fraction of the available samples. In our algorithm, picked samples are not removed from the pool of available training images, thus allowing the network to revisit certain training samples if required. This is especially crucial in the case of data imbalance since random selection has very low probability of selecting images from the down-sampled classes. Figure 4(D) shows the number of classification mistakes made by a CNN trained with our algorithm and with random selection. Classes 1, 3, 6, and 7 were significantly down-sampled. Our approach allows the CNN to perform well on these classes compared to random sampling.

## 3.2 VGG Face dataset

For the problem of face recognition, we choose to start with a pre-trained CNN in order to illustrate the use of our algorithm for transfer learning, as a fine-tuning sampling strategy. Using the methods and network described in [29], a CNN was pre-trained on the CASIA-WebFace dataset [30]. Instead of random initialization, we start with the pre-trained weights for the first 15 layers (up to the fifth pooling layer), and add two randomly initialized fully connected layers joined by a dropout layer. We train and test on the VGG Face dataset [31]. Since CASIA-WebFace and VGG Face have significant subject overlap, we choose 20 of the non-overlapping subjects. The VGG Face dataset consists of a large number of images, out of which a portion has been selected as part of the final curated set. We observe that the non-curated images are considerably more affected by label and bounding box noise. In order to get meaningful test results, we restrict our testing set to the curated images, while training on the entire dataset. We perform five-fold cross-validation using 5 random splits. We choose a budget of 100 training samples per loop of our algorithm.

Figure 5 shows some examples of images selected by our algorithm. The top images are selected in the first loop, when the representativeness score $\lambda_2$ is large and the uncertainty score $\lambda_3$ zero. We notice that all chosen samples are frontal, of good quality, and typical of the subjects. The bottom images are chosen much later in the process, after $\lambda_2$ has considerably decreased and $\lambda_3$ increased. This time, our method chooses more difficult examples which include extreme poses, obstruction, blur, an additional person, and an unusually young version of the subject.

We present the performance results of this CNN trained using our algorithm and random sampling in Table 1. Using only one loop (i.e., 100 picked images), the testing accuracy increases to $89.69\%$ compared to $80.05\%$ for random sampling. After 13 loops, the performance of random sampling plateaus at $94.89\%$, while our approach achieves $97.15\%$, which cuts the error in half.

| # of selected samples | 0 | 100 | 500 | 1300 | 1600 |
|---|---|---|---|---|---|
| Our approach | 12.73% | **89.69%** | **93.23%** | **97%** | **97.15%** |
| Random | 12.45% | 80.05% | 90.79% | 94.89% | 94.89% |

Table 1: VGG Face testing accuracies.

# 4 Conclusion

In this paper, we addressed the problem of reducing the training data requirement of deep neural networks. We proposed an efficient iterative and adaptive algorithm based on convex optimization. We demonstrated its effectiveness on real-life datasets and robustness to label noise and class imbalance.

**Acknowledgments**

This research is based upon work supported by the Office of the Director of National Intelligence (ODNI), Intelligence Advanced Research Projects Activity (IARPA), via IARPA R&D Contract No. 2014-14071600012. The views and conclusions contained herein are those of the authors and should not be interpreted as necessarily representing the official policies or endorsements, either expressed or implied, of the ODNI, IARPA, or the U.S. Government. The U.S. Government is authorized to reproduce and distribute reprints for Governmental purposes notwithstanding any copyright annotation thereon.

# DCNNs on a Diet: Sampling Strategies for Reducing the Training Set Size – Supplementary Material

Maya Kabkab, Azadeh Alavi, Rama Chellappa

Center for Automation Research, UMIACS
University of Maryland, College Park
{mayak, azadeh, rama}@umiacs.umd.edu

**Theorem 1** *The modified water-filling problem*

$$\max_{M_k^t \in \mathbb{Z}^+} \sum_{k=1}^{L} \log\left(1 + \alpha \cdot c_k^t \frac{M_k^t}{M^t}\right) \quad \text{s.t.} \quad \sum_{k=1}^{L} M_k^t \leq M^t; \ M_k^t \leq |\mathcal{X}_k|, \qquad (1)$$

*can be solved using Algorithm 1, as illustrated in Figure 1.*

---
**Algorithm 1** Integer water-filling algorithm with caps
---
1: Sort the base levels $\frac{M^t}{\alpha c_k^t}$ in ascending order and take the ceiling of the base levels $\lceil \frac{M^t}{\alpha c_k^t} \rceil$.
2: Place "caps" at $\lceil \frac{M^t}{\alpha c_k^t} \rceil + |\mathcal{X}_k|$.
3: **repeat**
4:     Fill with one water unit at a time proceeding from left to right without exceeding any cap.
5: **until** $M^t$ water units are used or all empty spaces are filled.
---

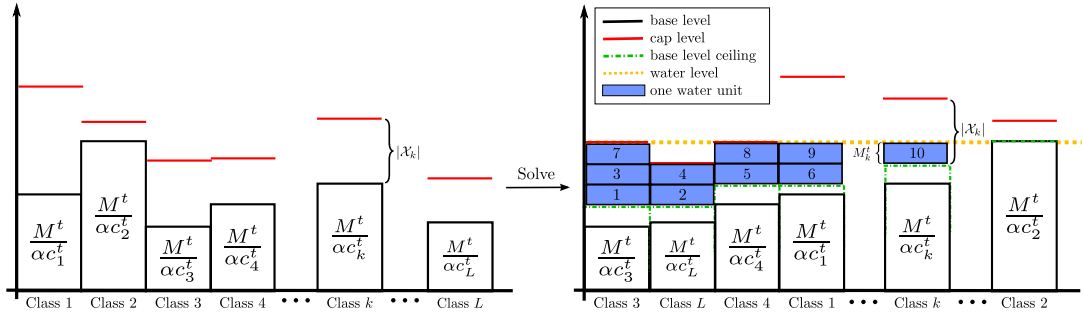

Figure 1: Integer water-filling with caps.

**Proof:** First, we ignore the $M_k^t \leq |\mathcal{X}_k|$ constraints, in (1), and let $h_k = \alpha c_k^t / M^t$. Then the problem in (1) becomes

$$\max_{M_k^t \in \mathbb{Z}^+} \sum_{k=1}^{L} \log\left(1 + h_k M_k^t\right) \quad \text{s.t.} \quad \sum_{k=1}^{L} M_k^t \leq M^t. \qquad (2)$$

We start by showing a necessary optimality condition.

**Lemma 1** *An optimal profile $\{M_1^{t\star}, M_2^{t\star}, \ldots, M_L^{t\star}\}$, must satisfy*

$$\left|\left(\frac{1}{h_i} + M_i^{t\star}\right) - \left(\frac{1}{h_j} + M_j^{t\star}\right)\right| \leq 1, \ \forall i, j \ such\ that\ M_i^t > 0, M_j^t > 0 \qquad (3)$$

.

**Proof of Lemma 1:** We prove the lemma by contradiction. Assume $\exists i, j$ such that

$$\left| \left( \frac{1}{h_i} + M_i^{t\star} \right) - \left( \frac{1}{h_j} + M_j^{t\star} \right) \right| = \Delta > 1 \tag{4}$$

Without loss of generality, assume that $\frac{1}{h_i} + M_i^{t\star} > \frac{1}{h_j} + M_j^{t\star}$. Now consider a new profile where $\overline{M_i^t} = M_i^{t\star} - 1$, $\overline{M_j^t} = M_j^{t\star} + 1$, and $\overline{M_k^t} = M_k^{t\star}$ for $k \neq i, j$. This new policy is clearly feasible. We consider the difference between the objective values achieved by $\{\overline{M_k^t}\}$ and $\{M_k^{t\star}\}$:

$$\sum_{k=1}^{L} \log\left(1 + h_k \overline{M_k^t}\right) - \sum_{k=1}^{L} \log\left(1 + h_k M_k^{t\star}\right)$$

$$= \log\left(1 + h_i \overline{M_i^t}\right) + \log\left(1 + h_j \overline{M_j^t}\right) - \log\left(1 + h_i M_i^{t\star}\right) - \log\left(1 + h_j M_j^{t\star}\right) \tag{5}$$

$$= \log \frac{\left(1 + h_i(M_i^{t\star} - 1)\right)\left(1 + h_j(M_j^{t\star} + 1)\right)}{\left(1 + h_i M_i^{t\star}\right)\left(1 + h_j M_j^{t\star}\right)} \tag{6}$$

$$= \log\left(1 + \frac{h_j(1 + h_i M_i^{t\star}) - h_i(1 + h_j M_j^{t\star}) - h_i h_j}{\left(1 + h_i M_i^{t\star}\right)\left(1 + h_j M_j^{t\star}\right)}\right) \tag{7}$$

$$= \log\left(1 + \frac{h_i h_j (M_i^{t\star} - M_j^{t\star} - 1) + h_j - h_i}{\left(1 + h_i M_i^{t\star}\right)\left(1 + h_j M_j^{t\star}\right)}\right) \tag{8}$$

$$\stackrel{(a)}{=} \log\left(1 + \frac{h_i h_j (\Delta - 1)}{\left(1 + h_i M_i^{t\star}\right)\left(1 + h_j M_j^{t\star}\right)}\right) \stackrel{(b)}{>} 0, \tag{9}$$

where (a) follows from the assumption in (4) and (b) from $\Delta > 1$. This shows that the profile $\{\overline{M_k^t}\}$ achieves a higher objective value than the profile $\{M_k^{t\star}\}$ which contradicts its optimality. ∎

Now, we show that Algorithm 1 solves the problem in (2). We proceed by induction on the budget $M^t$:

- Base case, $M^t = 1$: It is trivial to show that in the optimal profile, one water unit will be assigned to the class with the lowest base level $\frac{M^t}{\alpha c_k^t}$, with ties broken arbitrarily.

- Induction step: Given an optimal profile $\{M_k^t(m)\}$ that solves (2) for $M^t = m$, we find the optimal profile $\{M_k^t(m+1)\}$ for $M^t = m + 1$. It can be seen that one water unit should be added to $\{M_k^t(m)\}$ because any other deviation from this profile will violate Lemma 1. Furthermore, using similar arguments as in Lemma 1, we show that this water unit should be placed at the class with the smallest $\frac{M^t}{\alpha c_k^t} + M_k^t(m)$, with ties broken by the ordering of $\frac{M^t}{\alpha c_k^t}$.

This corresponds exactly to the operation of Algorithm 1 in the case of no caps. When caps are added, i.e., with the constraints $M_k^t \leq |\mathcal{X}_k|$, it is easy to show the following:

- If the optimal profile for Problem (2) is feasible for Problem (1), then it is also optimal for Problem (1).

- If the optimal profile for Problem (2) contains a class $i$ such that $M_i^t > |\mathcal{X}_i|$ (infeasible for Problem (1)), then the solution of Problem (1) must have $M_i^t = |\mathcal{X}_i|$. Furthermore, since no more water units can be allocated to this class, we can solve Problem (1) with class $i$ removed and total budget decreased by $|\mathcal{X}_i|$. This is equivalent to skipping class $i$ when its water level reaches its cap.

As this describes the operation of Algorithm 1 with caps, this proves the theorem. ∎

**Remark** The use of the ceiling operation $\lceil \frac{M^t}{\alpha c_k^t} \rceil$ in Algorithm 1 is not necessary for the optimality of the solution. However, it makes the procedure easier to visualize and results in the same optimal profile.

**Theorem 2** *The solution* $\mathbf{x}^*$ *to*

$$\min_{\mathbf{x} \in \{-1,1\}^N} \mathbf{x}^\mathsf{T} \mathbf{A}\mathbf{x} + \mathbf{b}^\mathsf{T}\mathbf{x} \quad \text{s.t.} \quad (\mathbf{1}^\mathsf{T}\mathbf{x} - 2M + N)^2 = 0. \tag{10}$$

*can be well-approximated by*

$$\hat{\mathbf{x}}^* = -\frac{1}{2}\left(\mathbf{A} + \mu^*\mathbf{1}\mathbf{1}^\mathsf{T} + \gamma^*\mathbf{I}\right)^\dagger (\mathbf{b} - 2\mu^*(2M - N)\mathbf{1}), \tag{11}$$

*where* $(\cdot)^\dagger$ *denotes the pseudo-inverse,* $\mathbf{I}$ *denotes the identity matrix in* $\mathbb{R}^{N \times N}$, *and* $\mu^*, \gamma^*$ *are the solution to the following semi-definite program (SDP):*

$$\max_{\mu,\gamma,\tau \in \mathbb{R}} \quad (2M - N)^2 \mu - \gamma N - \tau$$

$$\text{s.t.} \quad \begin{pmatrix} \tau & \frac{1}{2}(\mathbf{b} - 2\mu(2M - N)\mathbf{1})^\mathsf{T} \\ \frac{1}{2}(\mathbf{b} - 2\mu(2M - N)\mathbf{1}) & \mathbf{A} + \mu\mathbf{1}\mathbf{1}^\mathsf{T} + \gamma\mathbf{I} \end{pmatrix} \succeq 0 \tag{12}$$

**Proof:** We denote the optimal objective value of (10) by $p^\star$. We start by writing the Lagrangian dual. This problem has zero duality gap [1, Theorem 9], therefore:

$$p^\star = \max_{\mu \in \mathbb{R}} \min_{\mathbf{x} \in \{-1,1\}^N} \mathbf{x}^\mathsf{T}\mathbf{A}\mathbf{x} + \mathbf{b}^\mathsf{T}\mathbf{x} + \mu(\mathbf{1}^\mathsf{T}\mathbf{x} - 2M + N)^\mathsf{T}(\mathbf{1}^\mathsf{T}\mathbf{x} - 2M + N) \tag{13}$$

$$= \max_{\mu \in \mathbb{R}} \min_{\mathbf{x} \in \{-1,1\}^N} \mathbf{x}^\mathsf{T}\mathbf{A}\mathbf{x} + \mathbf{b}^\mathsf{T}\mathbf{x} + \mu\mathbf{x}^\mathsf{T}\mathbf{1}\mathbf{1}^\mathsf{T}\mathbf{x} - 2\mu(2M - N)\mathbf{1}^\mathsf{T}\mathbf{x} + \mu(2M - N)^2 \tag{14}$$

$$= \max_{\mu \in \mathbb{R}} \left\{ (2M - N)^2 \mu + \min_{\mathbf{x} \in \{-1,1\}^N} \mathbf{x}^\mathsf{T}\left(\mathbf{A} + \mu\mathbf{1}\mathbf{1}^\mathsf{T}\right)\mathbf{x} + (\mathbf{b} - 2\mu(2M - N)\mathbf{1})^\mathsf{T}\mathbf{x} \right\}. \tag{15}$$

Next, we relax the constraint $\mathbf{x} \in \{-1, 1\}^N$ to the non-binary constraint $\mathbf{x}^\mathsf{T}\mathbf{x} = N$. Commonly used relaxations for $\mathbf{x} \in \{-1, 1\}^N$ include $\mathbf{x}^\mathsf{T}\mathbf{x} = N$ and $x_i \in [-1, 1]$, resulting in the **same** duality gap [1, Theorem 2]. We use the former as it results another well-known problem with a zero-duality gap: the trust-region problem. Thus, we have:

$$p^\star \geq \max_{\mu \in \mathbb{R}} \left\{ (2M - N)^2 \mu + \min_{\mathbf{x}^\mathsf{T}\mathbf{x} = N} \mathbf{x}^\mathsf{T}\left(\mathbf{A} + \mu\mathbf{1}\mathbf{1}^\mathsf{T}\right)\mathbf{x} + (\mathbf{b} - 2\mu(2M - N)\mathbf{1})^\mathsf{T}\mathbf{x} \right\} \tag{16}$$

The inner minimization over $\mathbf{x}$ is known as the trust-region problem and has no duality gap [2, p.229], therefore:

$$(16) = \max_{\mu \in \mathbb{R}} \left\{ (2M - N)^2 \mu + \max_{\gamma \in \mathbb{R}} \left\{ \min_{\mathbf{x}} \mathbf{x}^\mathsf{T}\left(\mathbf{A} + \mu\mathbf{1}\mathbf{1}^\mathsf{T}\right)\mathbf{x} + (\mathbf{b} - 2\mu(2M - N)\mathbf{1})^\mathsf{T}\mathbf{x} + \gamma(\mathbf{x}^\mathsf{T}\mathbf{x} - N) \right\} \right\} \tag{17}$$

$$= \max_{\mu \in \mathbb{R}} \left\{ (2M - N)^2 \mu + \max_{\gamma \in \mathbb{R}} \left\{ -\gamma N + \min_{\mathbf{x}} \mathbf{x}^\mathsf{T}\left(\mathbf{A} + \mu\mathbf{1}\mathbf{1}^\mathsf{T} + \gamma\mathbf{I}\right)\mathbf{x} + (\mathbf{b} - 2\mu(2M - N)\mathbf{1})^\mathsf{T}\mathbf{x} \right\} \right\}$$

$$= \max_{\mu \in \mathbb{R}, \gamma \in \mathbb{R}} (2M - N)^2 \mu - \gamma N + \min_{\mathbf{x}} \left\{ \mathbf{x}^\mathsf{T}\left(A + \mu\mathbf{1}\mathbf{1}^\mathsf{T} + \gamma\mathbf{I}\right)\mathbf{x} + (\mathbf{b} - 2\mu(2M - N)\mathbf{1})^\mathsf{T}\mathbf{x} \right\} \tag{18}$$

We change the inner minimization over $\mathbf{x}$ to an equivalent form and rewrite (18) as:

$$(18) = \max_{\mu \in \mathbb{R}, \gamma \in \mathbb{R}} (2M - N)^2 \mu - \gamma N + \max_{\tau} -\tau \tag{19}$$

$$\text{s.t.} \quad \mathbf{x}^\mathsf{T}\left(A + \mu\mathbf{1}\mathbf{1}^\mathsf{T} + \gamma\mathbf{I}\right)\mathbf{x} + (\mathbf{b} - 2\mu(2M - N)\mathbf{1})^\mathsf{T}\mathbf{x} \geq -\tau \quad \forall \mathbf{x} \tag{20}$$

The constraint in (20) is equivalent to (from [3] and [4, Section 3.4]):

$$\begin{pmatrix} \tau & \frac{1}{2}(\mathbf{b} - 2\mu(2M - N)\mathbf{1})^\mathsf{T} \\ \frac{1}{2}(\mathbf{b} - 2\mu(2M - N)\mathbf{1}) & A + \mu\mathbf{1}\mathbf{1}^\mathsf{T} + \gamma\mathbf{I} \end{pmatrix} \succeq 0. \tag{21}$$

This concludes the proof. ∎

**Remarks**

1. We note that in the proof of Theorem 2, only one relaxation, (16), was made. Other relaxations are possible such as introducing $x_i^2 = 1 \ \forall i$ constraints and taking the Lagrangian dual as done in [5]. However, this results in $(N-1)$ additional optimization variables and makes the resulting SDP more complex. We do not adopt such an approach as we would like the subset selection problem to run as efficiently as possible.

2. The inner minimization over $\mathbf{x}$ in (18) is unconstrained and can be solved exactly, with a minimum value of
$$-\frac{1}{4}(\mathbf{b} - 2\mu(2M-N)\mathbf{1})^\intercal (A + \mu \mathbf{1}\mathbf{1}^\intercal + \gamma \mathbf{I})^\dagger (\mathbf{b} - 2\mu(2M-N)\mathbf{1}), \qquad (22)$$
if $(A + \mu \mathbf{1}\mathbf{1}^\intercal + \gamma \mathbf{I}) \succeq 0$. However, plugging the expression in (22) into (18) and maximizing over $\mu, \gamma$ does not result in a convex formulation and cannot be solved efficiently.